\documentclass[runningheads]{llncs}

\usepackage{eccv}

\usepackage{eccvabbrv}

\usepackage{graphicx}
\usepackage{booktabs}

\usepackage[accsupp]{axessibility}  

\usepackage{makecell}

\usepackage{wrapfig}

\usepackage[pagebackref,breaklinks,colorlinks]{hyperref}
\usepackage{hyperref}

\usepackage{orcidlink}

\usepackage{subcaption}

\usepackage[ruled,vlined]{algorithm2e}
\SetAlFnt{\small}

\title{Ada-adapter:Fast Few-shot Style Personlization of Diffusion Model with Pre-trained Image Encoder}
\titlerunning{Ada-adapter:Fast Few-shot Style Personlization of Diffusion Model}
\authorrunning{Jia Liu et al.}

\author{Jia Liu\inst{1}, Changlin Li\inst{2}, Qirui Sun\inst{2}, Jiahui Ming\inst{3}, Chen Fang\inst{2}, Jue Wang\inst{2}, Bing Zeng\inst{1} and Shuaicheng Liu\inst{1}*
}

\institute{
 University of Electronic Science and Technology of China, Chengdu, China\and
SeeKoo, Beijing, China\and
University of Chinese Academy of Sciences, Beijing, China
}

\begin{document}

\maketitle

\begin{abstract}
   Fine-tuning advanced diffusion models for high-quality image stylization usually requires large training datasets and substantial computational resources, hindering their practical applicability. We propose Ada-Adapter, a novel framework for few-shot style personalization of diffusion models. Ada-Adapter leverages off-the-shelf diffusion models and pre-trained image feature encoders to learn a compact style representation from a limited set of source images. Our method enables efficient zero-shot style transfer utilizing a single reference image. Furthermore, with a small number of source images (three to five are sufficient) and a few minutes of fine-tuning, our method can capture intricate style details and conceptual characteristics, generating high-fidelity stylized images that align well with the provided text prompts. We demonstrate the effectiveness of our approach on various artistic styles, including flat art, 3D rendering, and logo design. Our experimental results show that Ada-Adapter outperforms existing zero-shot and few-shot stylization methods in terms of output quality, diversity, and training efficiency.
   
\end{abstract}

\vspace{-0.1in}
\section{Introduction}
\label{sec:intro}

\begin{figure}[t]
\vspace{-0.1in}
  \centering
  \includegraphics[width=0.9\linewidth]{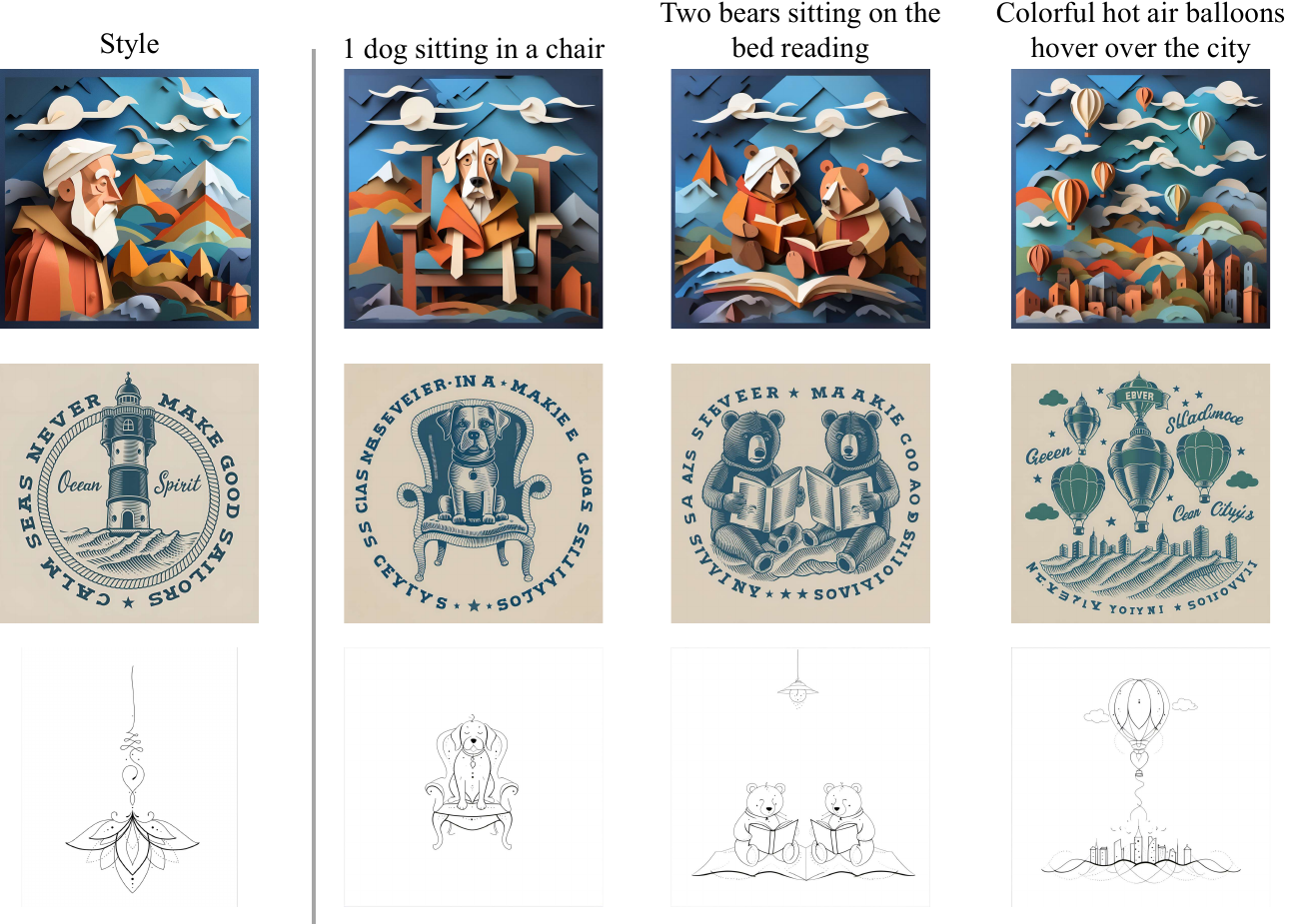}
  \caption{Results of our method for one-shot style transfer within 100 training steps and several minutes of fine-tuning.}
   \label{fig:ablation_one_shot}
    \vspace{-0.1in}   
\end{figure}
 
Generating realistic and other stylized images from natural language descriptions is a long-standing goal in artificial intelligence, but achieving it constantly and with low computational cost remains an open challenge. Recent advances in image generation based on large-scale text-to-image diffusion models, such as GLIDE~\cite{GLIDE}, IMAGEN~\cite{IMAGEN}, Stable Diffusion~\cite{StableDiffusion}, DALLE~\cite{Dalle1,Dalle2}, and others, have greatly stimulated the development of the art industry. These models can produce high-quality images within seconds from natural language descriptions, using a diffusion-based framework~\cite{ho2020denoising} that iteratively denoises a random noise image until it matches the desired text. However, generating images in a coherent style remains challenging.

\begin{figure}[t]
   \centering
   \includegraphics[width=1.0\linewidth]{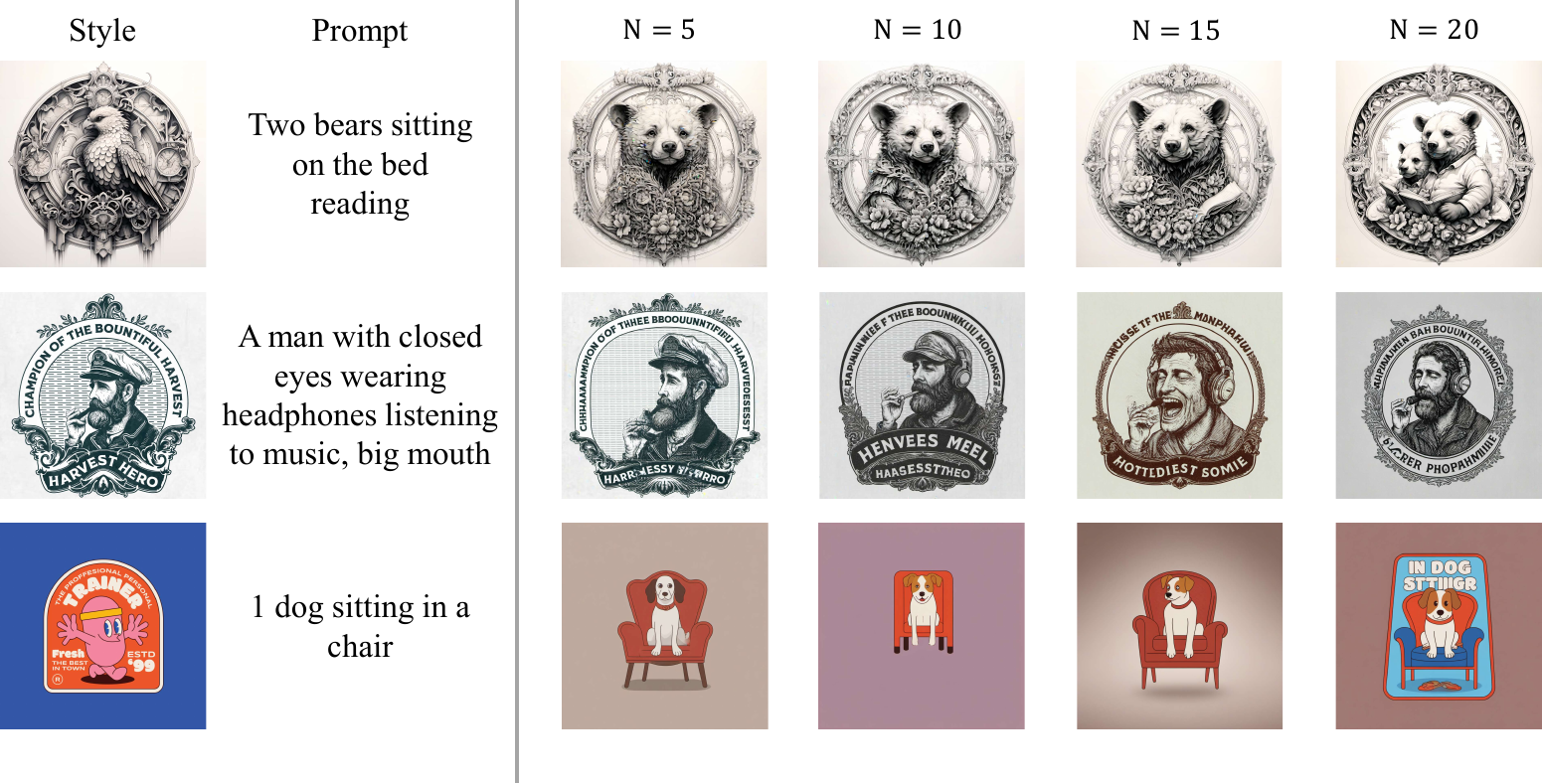}
   \caption{The stylization result of LoRAs trained with datasets of different sizes. We demonstrate the deterioration of the stylization quality and text alignment ability when the number of images $N$ ranges from 5 to 20 for training a style LoRA. The style generalization and text alignment ability becomes worse when fewer images are used. 
   }
   \label{fig:intro}
    \vspace{-0.2in}
\end{figure}

Existing approaches to address this challenge can be broadly classified into two categories: zero-shot methods and few-shot methods, each with its inherent limitations. On the one hand, zero-shot methods do not need test-time fine-tuning, and typically use a set of reference images to extract and inject style features into the denoising process. Some methods use a pre-trained image feature encoder to inject additional image features, such as ControlNet~\cite{zhang2023adding}, StyleAdaper~\cite{wang2023styleadapter} and IP-Adapter~\cite{ye2023ipadapter}. Other methods~\cite{hertz2024style,huang2024creativesynth,cao2023masactrl,ahn2023dreamstyler} share features within attention layers to transfer style from reference images. These zero-shot methods suffer from poor generalization ability and stability due to the limited capabilities of their pre-trained feature extractors or uncontrollable attention operations. On the other hand, few-shot stylization methods can incorporate extra knowledge into existing models, such as DreamBooth~\cite{ruiz2023dreambooth}, LoRA~\cite{hu2021LoRA}, Textual Inversion~\cite{gal2022image}, and other fine-tuning methods~\cite{lu2023specialist}, which can adapt a pre-trained diffusion model to any desired art style, but these methods still require dozens to hundreds of images, that share a consistent art style, to obtain a decent training result. Moreover, the quality of stylization degrades when the training data is insufficient. For example, as shown in Fig.~\ref{fig:intro}, we train vanilla LoRAs with the same 800 training steps but different sizes of datasets. The result shows that LoRA fails to achieve stable style generalization and satisfactory text alignment ability with a small amount of training data.


To address these issues, we propose a novel, simple, and efficient framework called Ada-Adapter, which is inspired by recent multi-modal conditioning methods, such as ControlNet~\cite{zhang2023adding}, T2I-Adapter~\cite{mou2023t2iadapter}, StyleAdapter~\cite{wang2023styleadapter}, BLIP-Diffusion~\cite{li2023blipdiffusion}, and IP-Adapter~\cite{ye2023ipadapter}, that use texts and images as conditions to guide the image generation. Our method can rapidly adapt existing diffusion models to a specific art style within less than three minutes of fine-tuning on a single RTX 4090, using only a few source images. For instance, our method can generate high-quality stylized images with only 1 image for training and the results are shown in Fig.~\ref{fig:ablation_one_shot}. 

Ada-Adapter combines image and text conditions to guide the fine-tuning and image generation process of the diffusion model. Specifically, we adopt pre-trained image feature encoders from IP-Adapter~\cite{ye2023ipadapter}, to extract features from reference images. We use the average of the extracted image features, to emphasize the style component in the image features, while weakening the individual subject feature of each image, which greatly reduces the training cost and the size of the training data. Moreover, we propose a layer-wise hierarchical strategy to improve the integration of image features, and to ensure the capability for both stylization and text alignment. The hierarchical strategy is able to directly convert a pre-trained image encoder from IP-Adapter~\cite{ye2023ipadapter} into a zero-shot style feature extractor. We conduct extensive experiments to explore the limitations of existing methods, and the advantages of our method. Our method overcomes the key challenge of unstable stylization and eliminates the need for massive training data in existing few-shot stylization methods.
We briefly summarize our contributions as follows:
\begin{itemize}
\item We propose Ada-Adapter, a method that expands existing few-shot methods to multi-modal condition for style personalization, achieving better stylization while requiring less training data and training time.
\item We propopse a simple but quite effective method that disentangles style and content from image condition, achieving high-quality zero-shot style transfer.
\item We conduct comprehensive experiments with various styles. The results shows that our method is superior to other few-shot methods in terms of effectiveness, stability, generalization and training cost. Our method is able to perform satisfactory and stable stylization requiring only 3 to 5 images for less than 5 minutes of fine-tuning on a single RTX 4090. We claim that our method achieves the state-of-the-art performance in the task of few-shot style personalization.

\end{itemize}
\vspace{-0.1in}
\section{Related Works}
\label{sec:related works}
\subsection{Text to image diffusion models}

Diffusion models~\cite{ho2020denoising} are a type of generative model that can create high-quality images from random noise by iteratively applying denoising operations. Recently, diffusion models have been extended to condition on text prompts~\cite{Dalle1,Dalle2,IMAGEN,StableDiffusion,podell2023sdxl,GLIDE}, enabling text-to-image synthesis with high fidelity and diversity. 
Stable Diffusion~\cite{StableDiffusion} uses a CLIP~\cite{radford2021learning} encoder to encode text into embeddings, and a diffusion model to generate images from text and noise.

These text-to-image diffusion models have demonstrated the power of combining large-scale language models and diffusion models for image synthesis, but they also face some challenges, such as the computational cost and complexity of the diffusion process, the trade-off between fidelity and diversity, and the generalization to different styles and domains.

\subsection{Diffusion-based stylization}
Image stylization is the task of applying a specific style to an image, such as artistic, cartoon, or sketch style. Diffusion-based stylization techniques leverage textual descriptions or reference images to enable versatile style manipulation and capture intricate style characteristics. In this section, we briefly review some recent diffusion-based stylization methods:
LoRA~\cite{hu2021LoRA} uses a low-rank adaptation technique to fine-tune large language models  with fewer parameters and better results. Textual Inversion~\cite{gal2022image} insert trainable parameters into text encoders with token embeddings, and is able to customize diffusion model with a few images. DreamBooth~\cite{ruiz2023dreambooth} also adds a unique identifier into text encoder and uses small amount of images to fine-tune diffusion models, and synthesizes novel images of the training concept.
ProSpect~\cite{zhang2023prospect} extends Textual Inversion embeddings to the time dimension, which uses token embeddings for diffferent time steps to represent, generate, and edit images with seperate attributes.
InST~\cite{zhang2023inversionbased} learns and transfers the artistic style of a painting directly from a single image, without complex textual descriptions, using an inversion-based style transfer method.
Diffusion in Style~\cite{everaert2023diffusion_in_style} take advantage of the distribution of style datasets and sample random latent tensors from target distribution to accelerate the fine-tuning process of a pre-trained diffusion model.
Style-Aligned~\cite{hertz2024style} put forwards a zero-shot methods which share intermediate variables across attention layers to perform a consistent stylization.
StyleDrop~\cite{sohn2023styledrop} takes advantage of iterative training to adapt diffusion models to any desired style, but the manual intervention requires extra efforts and may lead to unstable performance.
These methods have shown the effectiveness and flexibility of diffusion-based stylization, but they struggles to meet the need of both style and text alignments with limited style references. Our method uses off-the-shell image encoders to disentangle and inject style from reference images, and is able to perform stable stylization with strong ability of generalization and text understanding, requiring only 3 to 5 source images and a few minutes of fine-tuning.

\vspace{-0.1in}
\section{Method}

\begin{figure*}[t]
   \centering
   \includegraphics[width=1.0\linewidth]{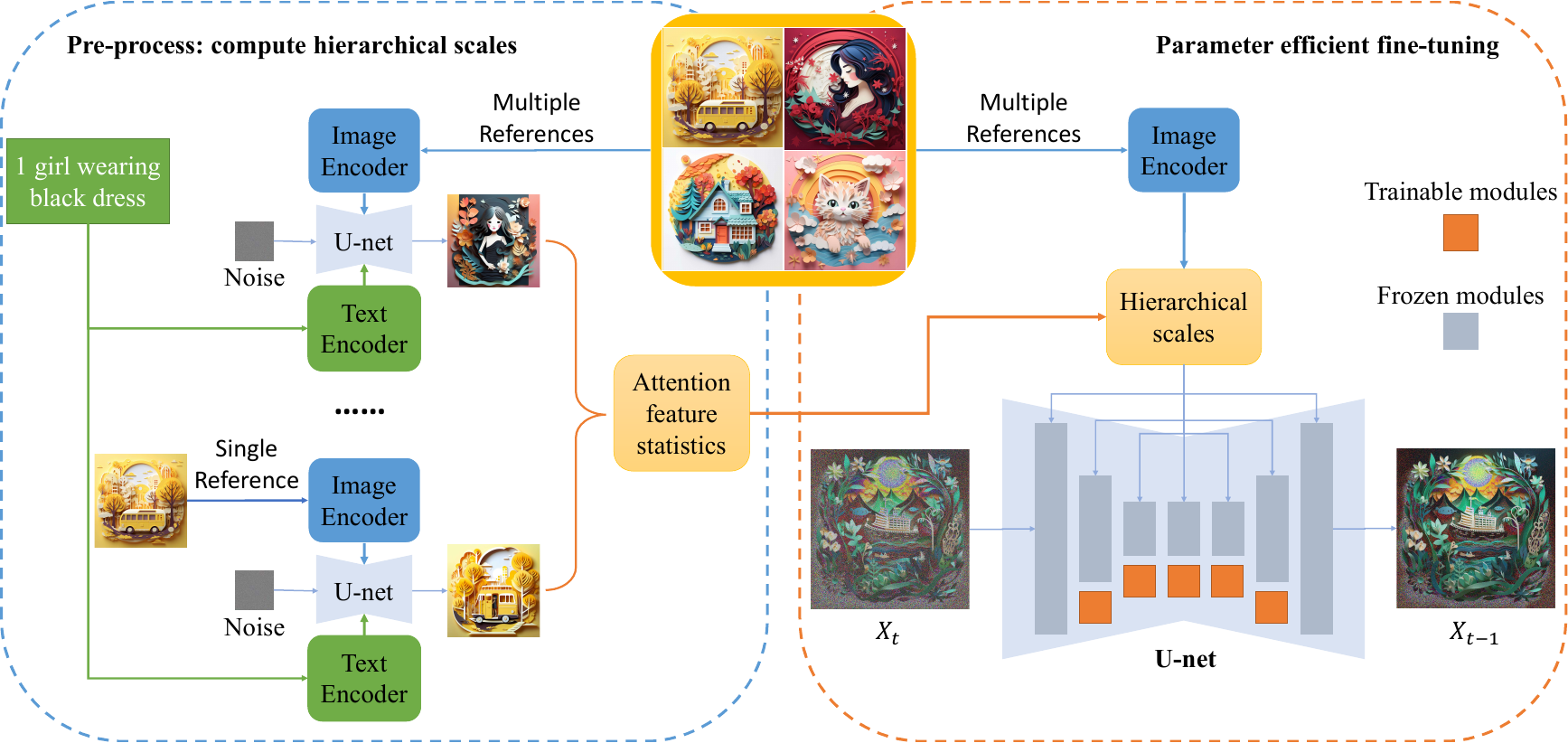}\\
   \caption{An overview of our method pipeline. We use a pre-trained image encoder to enable multi-modal denoising. Our pipeline consists of two main parts. The first part is on the left of the figure, where we perform inference processes with the fixed image encoder and different reference inputs. We record the intermediate attention features to compute the hierarchical scales for the image encoder. The hierarchical scales are applied to the image encoder to better disentangle style and subject features. The second part is on the right of the figure, where we fine-tune the diffusion model using LoRA modules with image conditions.
}
   \label{fig:pipeline}
\vspace{-0.2in}
\end{figure*}

In this section, we introduce the preliminaries of our method, including diffusion model~\cite{ho2020denoising, StableDiffusion} and IP-Adapter~\cite{ye2023ipadapter}. We illustrate the problem existing in current stylization methods based on diffusion models. Then we propose Ada-Adapter, a noval few-shot framework for efficiently adapting a pre-trained diffusion model to a style domain. We also put forth a key hierarchical strategy that fully activates the capabilities of our method and ensures satisfactory performance for both style and text alignment.

\vspace{-0.1in}
\subsection{Preliminaries: Diffusion model}
Diffusion models~\cite{ho2020denoising} are a set of probabilistic models trained to predict a distribution of $p(x_0)$ that approximates the real data distribution $q(x_0)$ by gradually denoising a random variable $x_t$ in $T$ time steps which fits standard Gaussian distribution $\mathcal{N}(0,I)$ at start point. 


Recent popular diffusion models~\cite{Dalle2,StableDiffusion,podell2023sdxl} employs a U-net based architecture to perform denoising tasks. The cross attention mechanism is introduced into U-net architectures, which enables these diffusion models to be conditioned by text. The objective function for training diffusion models is shown in Eq.~\ref{equation:objective_function}.

\vspace{-0.1in}
\begin{equation}
    L = \mathbb{E}_{x_0,\epsilon\sim\mathcal{N}(0,I),c_{t},t} \left \| \epsilon - \epsilon_{\theta}(x_t,c_{t},t)\right \|
    \label{equation:objective_function}
\end{equation}
where $\epsilon$ is gaussian noise, $c_{t}$ represents embeddings of text condition, $\theta$ denoted the trainable parameters of U-net, $t$ is time step and $t\in[0,1000]$, $x_0$ is a clean sample of datasets. The relationship between $x_0$ and $x_t$ is shown in Eq.~\ref{equation:forward_process}.
\begin{equation}
    x_t=\sqrt{\alpha_t}x_0+\sqrt{1-\alpha_t}\epsilon
    \label{equation:forward_process}
\end{equation}
where values of $\alpha_t$ are pre-defined values by a fixed schedule.

IP-Adapter~\cite{ye2023ipadapter} introduces image condition $c_{i}$ into diffusion models, which guides the denoising process by a reference image, and generate images that align with the reference image at main subjects and styles. IP-Adapter injects image features via cross attention layers in U-net which can be simplified in Eq.~\ref{equation:cross_attention}.

\begin{equation}
\begin{aligned}
    Z &= Z_{t}+  Z_{i}&=Softmax(\frac{QK^T}{\sqrt{d}})V+  Softmax(\frac{Q(K_i)^T}{\sqrt{d}})V_i 
\label{equation:cross_attention}
\end{aligned}
\end{equation}
where $Z$ is the final output feature of cross attention layer, $Z_{t}$ and $Z_{i}$ are features of text-conditioned features and image-conditioned features respectively. $Q=Z_{in} W_q, K=c_{t} W_k, V=c_{t} W_v$ are the query, key, value matrices of attention operation. On the other side, $K_i=c_{i} W_{ik}, V_i=c_{i} W_{iv}$ are key and value matrices of image conditions where $c_{i}$ is the image condition embedding and $W_{ik}, W_{iv}$ are projection matrices in cross attention layers of IP-Adapter.

\begin{figure*}[t]
   \centering
   \includegraphics[width=1.0\linewidth]{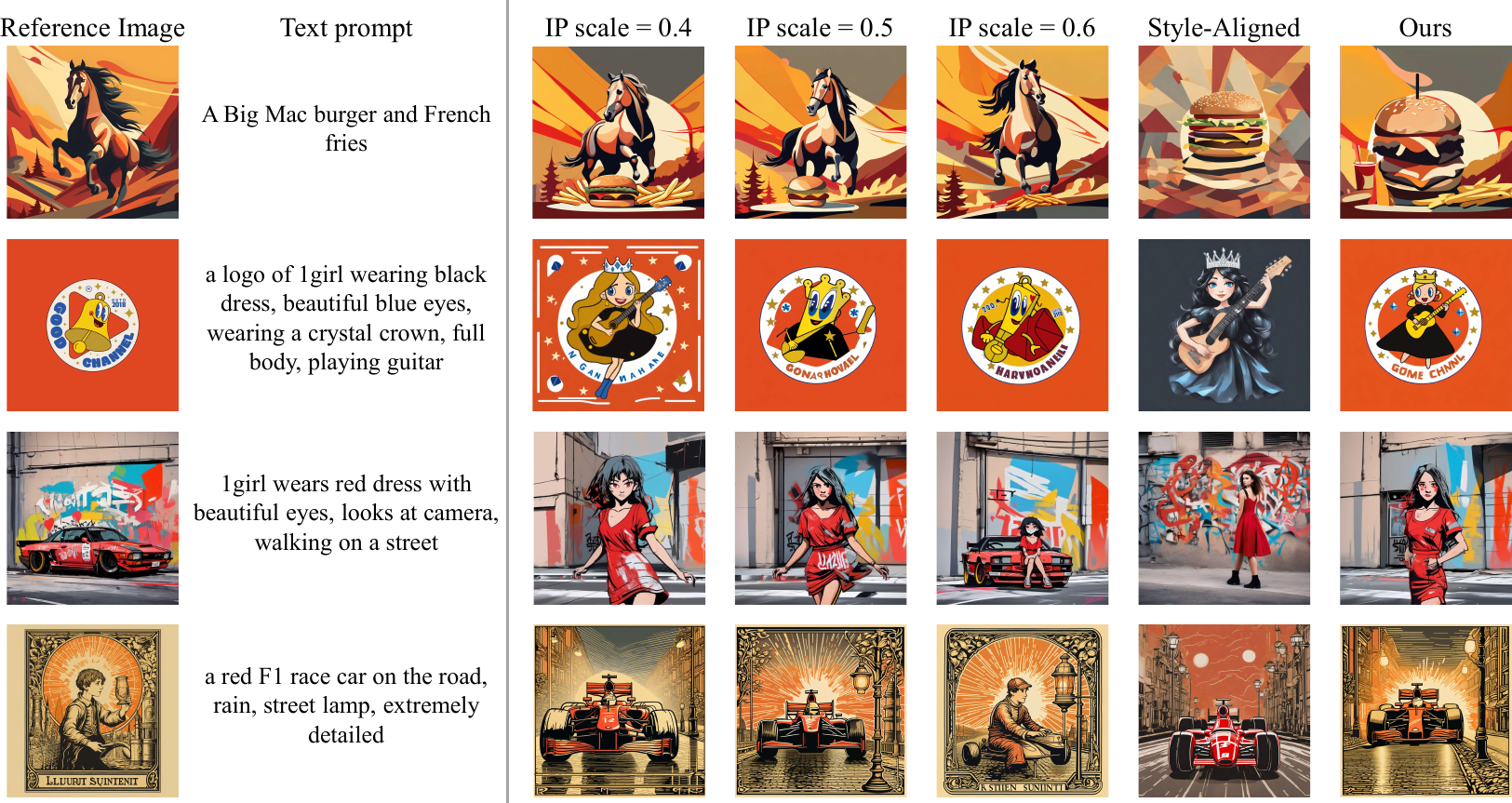}\\
   \caption{The effectiveness of hierarchical scales. We perform denoising process based on SDXL~\cite{podell2023sdxl}. On the left of this figure are input reference images and text prompts, and on the right of the figure are zero-shot stylized results. We set IP-Adapter with various scales, and to be specific, when the scale directly is set to $\lambda$, the output features of all layers in IP-Adapter are scaled by $\lambda$. In contrast, our method assigns unique scales to individual layers, effectively preserving the stylistic integrity of the reference images while concurrently mitigating semantic discrepancies.
   }
   \label{fig:zeroshot_hierarchical_weights}
\vspace{-0.1in}
\end{figure*}

\subsection{Visual modality condition for stylization}
\label{sec:multi_modal_condition}


To mitigate the limited generalization capability in the existing methods, we leverage pre-trained image encoders from IP-Adapter~\cite{ye2023ipadapter} to incorporate additional visual modality features into the denoising process of the diffusion model.
To take advantage of the visual modality condition, as the main function of IP-Adapter is to extract general features from images and inject them into denoising, we compute image embeddings with the image encoder of IP-Adapter and make use of these embeddings to provide style prior from reference images.

Learning from the design of CLIP~\cite{radford2021learning}, we assume that the content of an image can be described as an $X$ subject in $Y$ style and the image embedding spaces are nearly linear spaces. For a single reference image $R$,  the image condition $c_{i}$ contains the subject information $c_{i}^{subject}$ that describes the main subjects and the style information $c_{i}^{style}$ that denotes the art style of the reference image, and in our hypothesis, $c_{i} = c_{i}^{style}+c_{i}^{subject}$. For a datasets that contains images of a consistent style, we figure the images share the same $c_{i}^{style}$ but different $c_{in}^{subject}$, where $n$ is the index of the reference image and $c_{in}$ is the image condition of the $n^{th}$ image. Naturally, as is shown in Eq.~\ref{equation:average_of_image_embedding}, by taking average of the multiple image embeddings, the subject information of individual images is degraded, and the style information of the reference images is kept or enhanced, which provides abundant prior style knowledge for the denoising process.
\begin{equation}
\centering
\begin{aligned}
&\lim_{N\to\infty }\frac{1}{N}\sum_{n=1}^{N}{c_{in}^{subject}}=0  \\
\lim_{N\to\infty }\frac{1}{N}\sum_{n=1}^{N}{c_{in}} &= \lim_{N\to\infty }\frac{1}{N}\sum_{n=1}^{N}c_{in}^{subject} + \lim_{N\to\infty }\frac{1}{N}\sum_{n=1}^{N}c_{in}^{style} \\ &=c_{i}^{style}
\end{aligned}
\vspace{-0.1in}
\label{equation:average_of_image_embedding}
\end{equation}
where $N$ is the number of reference images. Thus, we can inject style features into denoising process with IP-Adapter by simply using the average of image embeddings from references.

\subsection{Hierarchical Adapter}
\label{subsec:hierarchical adapter}

\begin{figure}[t]
\vspace{-0.2in}  
   \centering
   \includegraphics[width=0.9\linewidth]{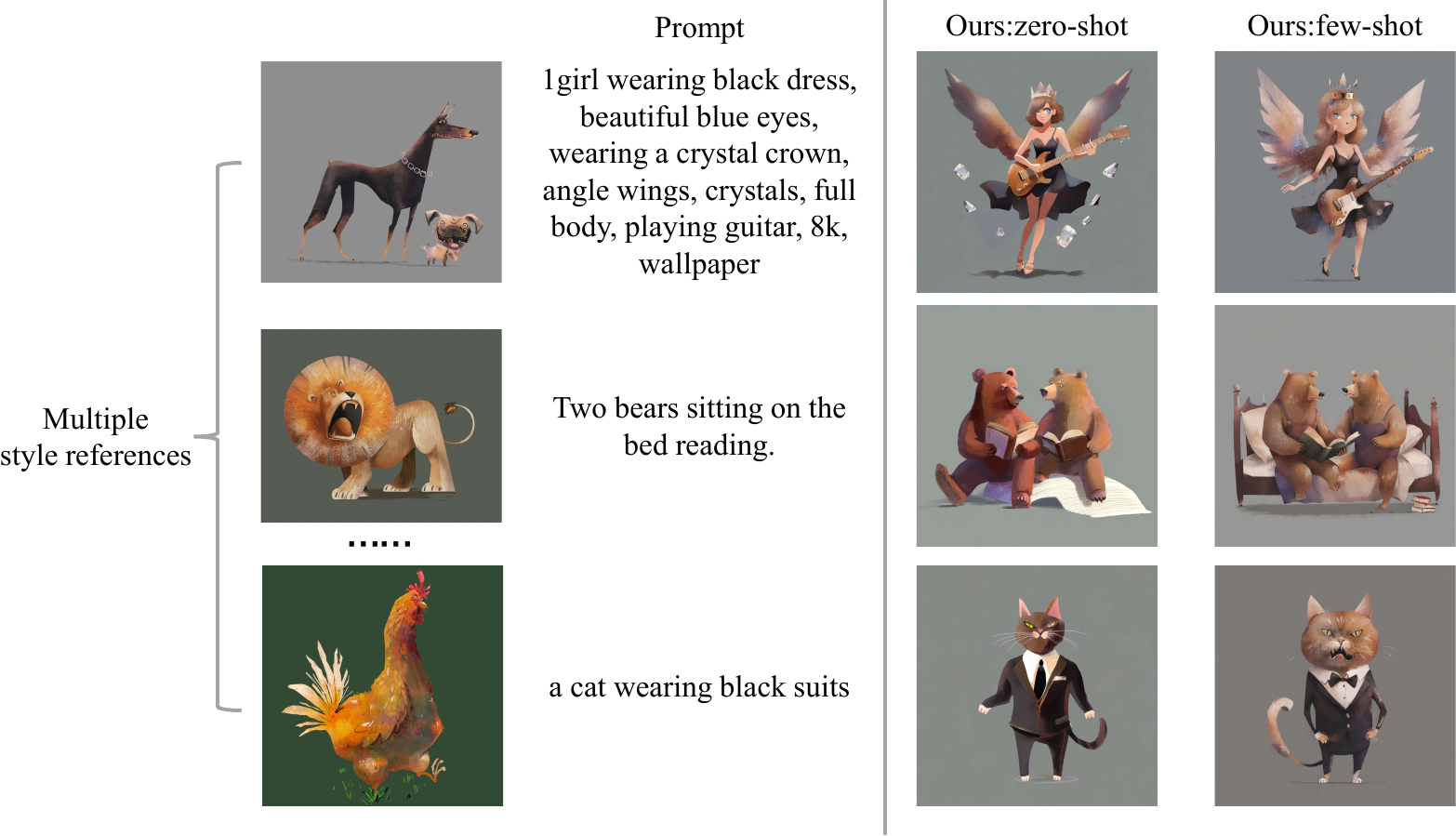}\\
   \caption{The outcomes of our method for both zero-shot and few-shot style transfer. The figure presents, on the left, three reference images that exemplify a flat, exaggerated character illustration style. On the right, we illustrate the results of our zero-shot method, which adeptly replicates the flat art style and textures. However, it does not fully preserve the exaggerated characteristics inherent to the style, a feat our few-shot method accomplishes with greater fidelity.
   }
   \label{fig:zero_shot_failure}
\vspace{-0.3in}   
\end{figure}
In previous section, we assume that we can inject style features into denoising process by using multiple reference images. However, in practice, even when we use the average image embedding as the condition, the limited number of reference images (sometimes only 1 image is provided) makes it difficult to completely eliminate the effect of $c_i^{subject}$ as described in Eq.~\ref{equation:average_of_image_embedding}. This adversely affects the text-alignment ability of the diffusion model. 

To investigate the influence of the image condition in the denoising process, we calculate the cosine similarities between the final output of the cross attention and two intermediate outputs: (1) the output of the text-conditioned cross attention, (2) the output of the image-conditioned cross attention. This calculation is performed for each cross attention layer and each time step:

\begin{equation}
\begin{aligned}
    \centering
    P_{t}(d,t) &=  cos(Z_{t}(d,t),Z(d,t))\\
    P_{i}(d,t) &= cos(Z_{i}(d,t),Z(d,t)) 
\end{aligned}
\label{equation:cosine}
\end{equation}
where $d,t$ are cross attention layer index and time step index, $P_{t}$ and $P_{i}$ are the contribution of text prompt and reference image measured with cosine similarity.

\begin{wrapfigure}{htr}{0.4\textwidth}
\vspace{-0.1in}
  \begin{center}
  \vspace{-0.4in}
       \includegraphics[width=1\linewidth]{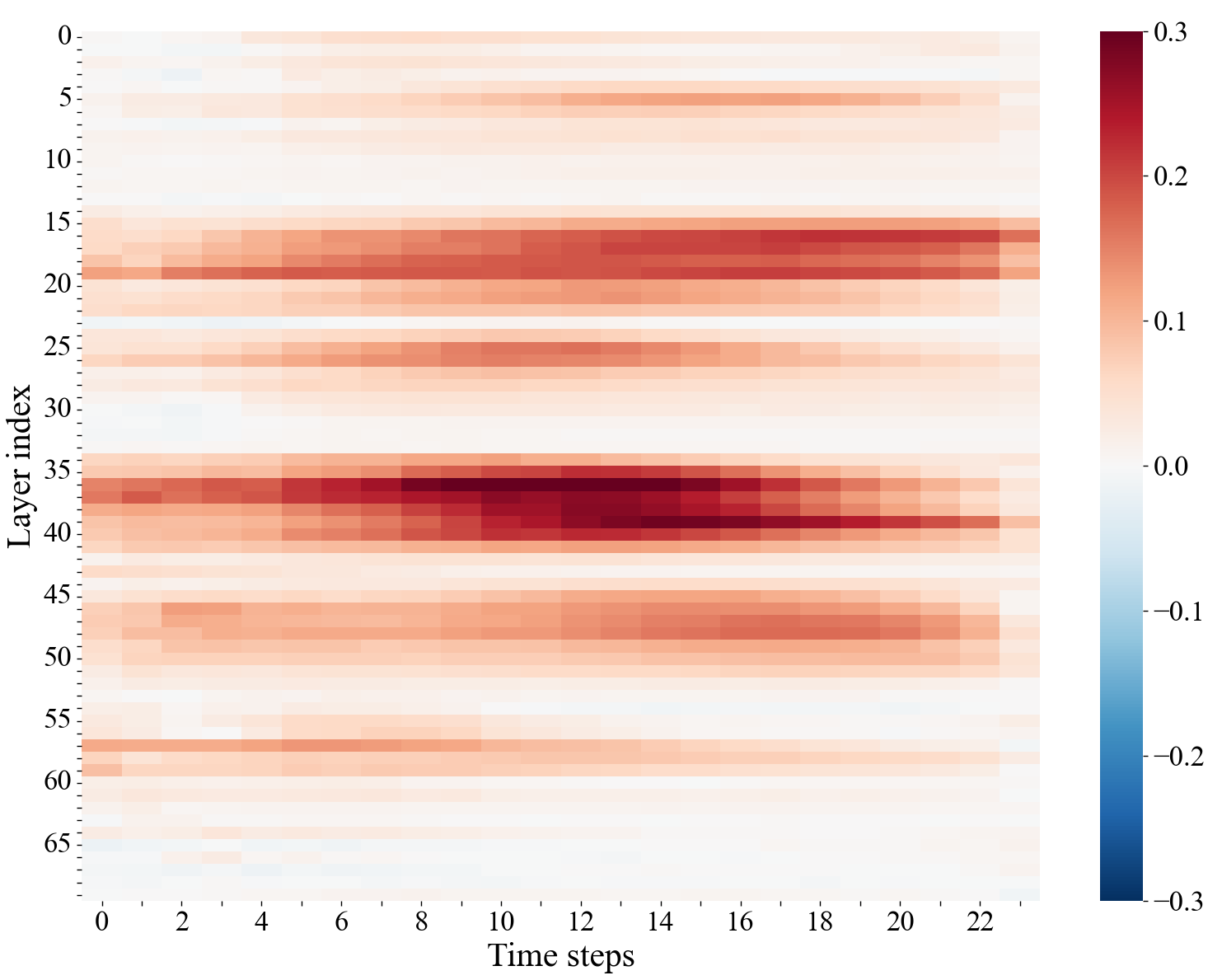}\\
  \end{center}
  \vspace{-0.2in}
     \caption{The results of $P_{i} - P_{t}$ across all time steps and layers. In most of the layers, image feature takes the lead, weakening the influence of text prompts.}
   \label{fig:cos difference}
   \vspace{-0.2in}
\end{wrapfigure}


In our analysis, depicted in Fig.~\ref{fig:cos difference}, we observe a predominant influence of the image condition over the text condition across most layers, potentially impairing the text alignment in the final outputs. 

According to the discoveries in~\cite{cao2023masactrl} and the practice of community~\cite{LoRABlockWeights}, diffusion models exhibit hierarchical characteristics, which means that certain layers of the U-net prioritize the main subject, while others focus on the artistic style and finer details. Drawing inspiration from this hierarchical structure, we propose that IP-Adapter also possesses similar attributes, with designated layers infusing more object-centric information from reference images and others channeling high-frequency style details.

Building on this hypothesis, as articulated in Eq.~\ref{equation:average_of_image_embedding}, employing the average of image embeddings as condition in the denoising process dilutes the sum of $c_{subject}$ compared to the use of a singular image embedding. Consequently, we compute $P_{i}$ across all layers and temporal steps during separate inferences with averaged image embeddings $c_{i}$ and individual image embeddings $c_{in}$:
\begin{equation}
\begin{aligned}
    \centering
    P_{i}^{n}(d,t) &=  cos(Z_{i}(d,t,c_{in}),Z(d,t,c_{in}))\\
    P_{i}^{M}(d,t) &= cos(Z_{i}(d,t,c_{i}),Z(d,t,c_{i})) 
\end{aligned}
\label{equation:cosine_image}
\end{equation}
where $P_{i}^{n}$ is the contribution of image feature conditioned by the $n^{th}$ image embedding $c_{in}$, $P_{i}^{M}$ is the contribution of image features conditioned by the average image embedding $c_i$, and $c_{i} = \frac{1}{N}(c_{i1}+c_{i2}+...+c_{i(N-1)}+c_{iN})$, $M$ refers to multiple reference images.


When employing the average image embedding as a condition, the cross attention layer $d$ that predominantly introduces subject information should have a diminished influence from the image condition compared to a single image embedding. This implies that  $|P_{i}^{M}| < |P_{i}^{n}|$ within this layer. To maintain the stylistic elements while minimizing subject features in the image condition, we have devise a straightforward strategy. The strategy involves scaling down the image-conditioned features in layers that predominantly transfer subject information, while preserving the scales in layers that convey style, and the strategy is summarized in the following equation:

\begin{equation}
\begin{aligned}
    scale &= \mathbb{S}(P_{i}^{n},P_{i}^{M},P_t^n,P_t^M), scale \in [0,1] \\
    Z &= Z_{t} + scale \times Z_{i}
\label{equation:strategy_scale}
\end{aligned}
\end{equation}

where $\mathbb{S}$ is a strategy that computes the $scale$ for image-conditioned features based on $P_{i}^{n}$, $P_{i}^{M}$ and $P_t^n,P_t^M$. Note that $P_t^n,P_t^M$ are contributions of text condition when using single or multiple reference images, similar to the definition in Eq.~\ref{equation:cosine_image} for image condition. We illustrate our strategy $\mathbb{S}$ in Algorithm~\ref{alg1:hierarchical_strategy}. In practice, we use $D(d,t)=P_{i}(d,t)-P_{t}(d,t)$ to better to measure the contribution of image condition, which takes the text condition as a base. $D^M(d)>D^S(d)$ means the layer transfers more global style features, while in contrast, the layer focus more on the individual feature of a single reference image. We set the scale of a layer $d$ according to the difference between $D^S(d)$ and $D^M(d)$.

\begin{algorithm}
\caption{The strategy for computing hierarchical scales}
\label{alg1:hierarchical_strategy}

\KwIn{Reference style images $R_1,R_2...R_N$, pre-trained diffusion model $\theta$, text prompt condition embedding $c_t$, pre-trained image encoder $V$ and attention processors from IP-Adapter, inference time steps $T$}
\BlankLine
\textbf{Initialization:} \\
Compute $c_{in}=V(R_n), n \in [0,1,2,...N]$\\
Compute $c_{i} = \frac{1}{N}(c_{i1}+c_{i2}+...+c_{i(N-1)}+c_{iN})$\\
Define multi-modal inference pipeline with IP-Adapter as $f_{\theta}$
\BlankLine
\For{$n$ in range($N$)}{
    \While{inference with single reference image $f_{\theta}(c_t,c_{in},T)$}{
        Compute $P_{i}^{n}(d,t)$ and $P_{t}^{n}(d,t)$ based on Eq.~\ref{equation:cosine} and Eq.~\ref{equation:cosine_image}\\
        Compute $D^n(d,t)=P_{i}^{n}(d,t)-P_{t}^{n}(d,t)$
    }
}
Compute $D^S(d)=\frac{1}{NT}\sum_{n=1}^{N}\sum_{t=1}^{T}D^n(d,t)$\\
\While{inference with multiple references $f_{\theta}(c_t,c_{i},T)$}{
    Compute $P_{i}^{M}(d,t)$ and $P_{t}^{M}(d,t)$ based on Eq.~\ref{equation:cosine} and Eq.~\ref{equation:cosine_image}\\
    Compute $D^M(d,t)=P_{i}^{M}(d,t)-P_{t}^{M}(d,t)$
}
Compute $D^M(d)=\frac{1}{T}\sum_{t=1}^{T}D^M(d,t)$\\
Compute $Scales(d) = Normalize(D^M(d)-D^S(d))$

\KwOut{$Scales$}  
    
\end{algorithm}

To illustrate the effectiveness of our hierarchical method, we show the result of zero-shot style transfer comparing with IP-Adapter~\cite{ye2023ipadapter} and Style-Aligned~\cite{hertz2024style} in Fig.~\ref{fig:zeroshot_hierarchical_weights}. The result indicates that our method of hierarchical scales helps to preserve more style prior knowledge while remain the ability of text alignment, which is superior to other methods.

\begin{figure*}[t]
   \centering
   \includegraphics[width=1.0\linewidth]{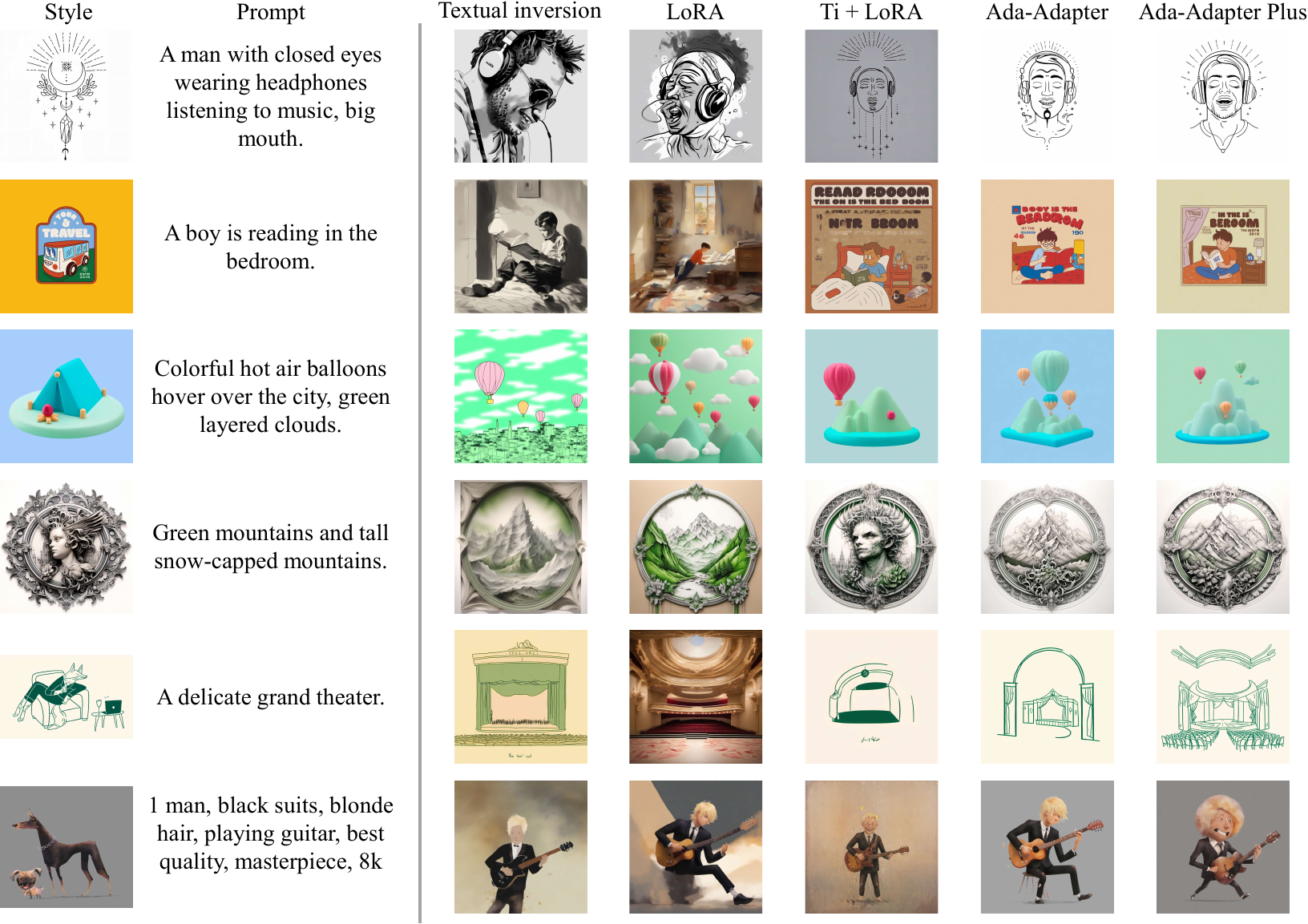}\\
   \caption{Qualitative comparisons with other few-shot methods.
   }
   \label{fig:exp_qualitative}
\vspace{-0.3in}   
\end{figure*}

\subsection{Few-shot style personalization}
\label{sec:few_shot_fine-tuning}

In the preceding section, we demonstrated that our hierarchical scale strategy effectively transforms a pre-trained image encoder into a zero-shot style feature extractor. However, this approach encounters limitations when dealing with styles not solely characterized by color, structure, or texture. Such styles’ intrinsic concepts and subtle details are often ignored by zero-shot methods, as evidenced in Fig.~\ref{fig:zero_shot_failure}, where our zero-shot method fails to capture the style’s exaggerated essence, resulting in characters with proportionate limbs.

Nonetheless, reference images remain a potent source of style information, serving as a valuable foundation for fine-tuning methods. This advantage is particularly useful in few-shot stylization, significantly reducing the reliance on extensive training datasets and shortening the training duration.

Therefore, we include image features into the fine-tuning process and extend the training objective function in Eq.~\ref{equation:objective_function} to multi-modal condition, the new objective function is shown in Eq.~\ref{equation:multi-modal fine-tuning}.

\begin{equation}
\centering
    L = \mathbb{E}_{x_0,\epsilon\sim\mathcal{N}(0,I),c_{t},c_{i},t} \left \| \epsilon - \epsilon_{\theta}(x_t,c_{t},c_{i},t)\right \| 
\label{equation:multi-modal fine-tuning}
\end{equation}
 We follow common practice ~\cite{hu2021LoRA} to add trainable low-rank modules in attention layers and keep the base model's parameters frozen. The overall multi-modal fine-tuning pipeline is named as Ada-Adapter and illustrated in Fig.~\ref{fig:pipeline}.

\vspace{-0.1in}

\section{Experiments and evaluations}

We implement our method and conduct experiments based on Stable Diffusion XL~\cite{podell2023sdxl}. To demonstrate the effectiveness of our method on few-shot stylization, we have collected 16 datasets of different styles to evaluate our method. Each dataset contains 5 source images, and images are captioned by large multi-modal language models~\cite{zhu2023minigpt4,touvron2023llama}. 

\subsection{Qualitative comparison}

\begin{table}[ht]
\vspace{-0.3in}
\centering
\resizebox{1.0\linewidth}{!}{
\begin{tabular}{p{2.5cm}<{\centering}|p{2.5cm}<{\centering}|p{2.5cm}<{\centering}|p{2.5cm}<{\centering}|p{2.5cm}<{\centering}|p{2.5cm}<{\centering}}
\toprule
            & TI  & LoRA & TI + LoRA  & Ada-Adapter & Ada-Adapter Plus \\
\midrule
Art Fid ($\downarrow$)        & 10.660  & 8.727 & 7.345 & \underline{5.958}   & \textbf{5.187}      \\                                        
Clip Score ($\uparrow$)        & 23.592  & \textbf{26.749} & 17.539 & \underline{24.533}   & 24.179     \\                                                          
\bottomrule
\end{tabular}
}
\caption{Quantitative comparison with other few-shot methods. We use bold font to represent the best performance, and underline refers to the second best result.}
\label{tab:comparison}
\vspace{-0.4in}
\end{table}
We assess the performance of our Ada-Adapter on collected style datasets, adhering to the best practices in community for training conventional Textual Inversion, LoRA, and the combined Textual Inversion + LoRA for comparative analysis. Across all methods, each image in a style dataset undergoes 100 iterations of training. Subsequently, we conduct inference using over 200 prompts from DrawBench~\cite{saharia2022photorealistic}, generating three distinct samples per prompt.

TI serves as an abbreviation for Textual Inversion. The hybrid approach, Textual Inversion + LoRA, means we first train a vanilla Textual Inversion embedding for the target style, then we use the same dataset to train a style LoRA, utilizing the pre-trained Textual Inversion token as a trigger word within the caption. This two-stage process obviates the necessity for manual LoRA trigger word configuration to suit the style.

Multiple image encoders from IP-Adapter~\cite{ye2023ipadapter} are available for Stable Diffusion XL~\cite{podell2023sdxl}. We demonstrate our method using IP-Adapter XL and IP-Adapter Plus XL as image encoders. Specifically, Ada-Adapter incorporates IP-Adapter XL, whereas Ada-Adapter Plus utilizes IP-Adapter Plus XL as its image encoder.

The qualitative evaluation, depicted in Fig.~\ref{fig:exp_qualitative}, reveals that both TI and LoRA alone are insufficient for producing satisfactory stylized outcomes with a mere five source images. Although TI + LoRA manages to grasp certain style elements, it falls short of harmonizing with the textual prompts. In contrast, our Ada-Adapter consistently generate images that align with both the semantic content of the prompts and stylistic essence from reference images.
\vspace{-0.1in}
\subsection{Quantitative evaluation}
Following the practice in ~\cite{everaert2023diffusion_in_style}, we use art-fid~\cite{wright2022artfid} and Clip Score~\cite{hessel2022clipscore} to evaluate the stylization and text alignment ability of our method and others. The results are shown in Table~\ref{tab:comparison}. 

We note that the vanilla LoRA achieves the highest Clip Score among all these methods, which we consider does not correspond to the real situation. The CLIP~\cite{hessel2022clipscore,radford2021learning} cannot disentangle style and content completely, which means the Clip Score will be lower if we compute it without descriptions about the art style, compared to the full description that includes the art style and the content. As in our experiments, the style datasets are processed and the descriptions about styles are removed manually, naturally we compute the Clip Score between the generated images and the text prompts that only describe the main subject, excluding the description about the art style which results in a lower Clip Score when stylization is better performed. We conduct more comparisons in Sec.~\ref{subsec:ab_scale_multiplier}.

\begin{figure}[htbp]
\vspace{-0.1in}
\centering
\begin{minipage}[t]{0.48\textwidth}
\centering
\includegraphics[width=5.5cm]{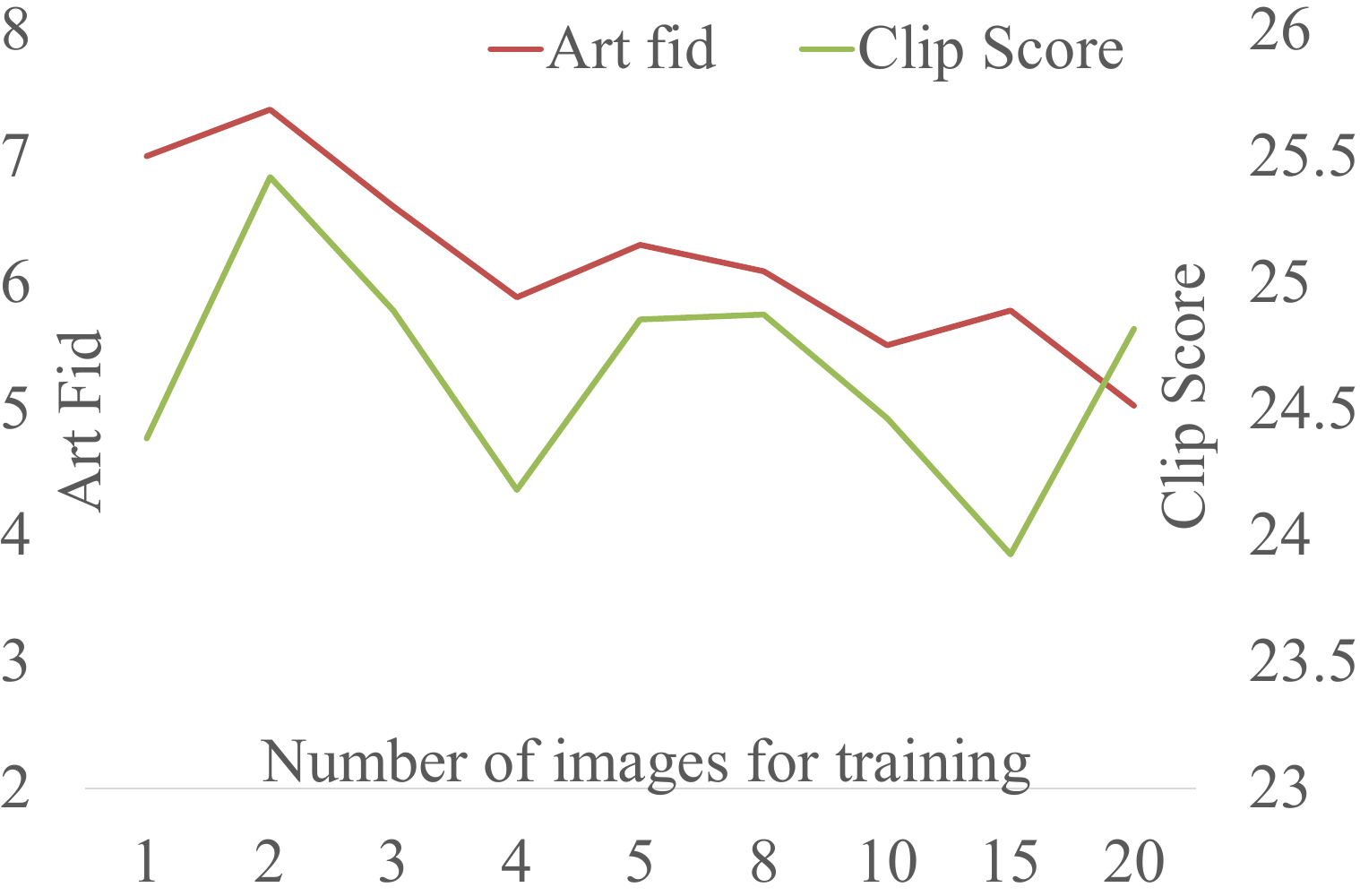}
\caption{Influence of dataset sizes.}
\label{fig:ablation_image_count}
\end{minipage}
\hfill
\begin{minipage}[t]{0.48\textwidth}
\centering
\includegraphics[width=5.5cm]{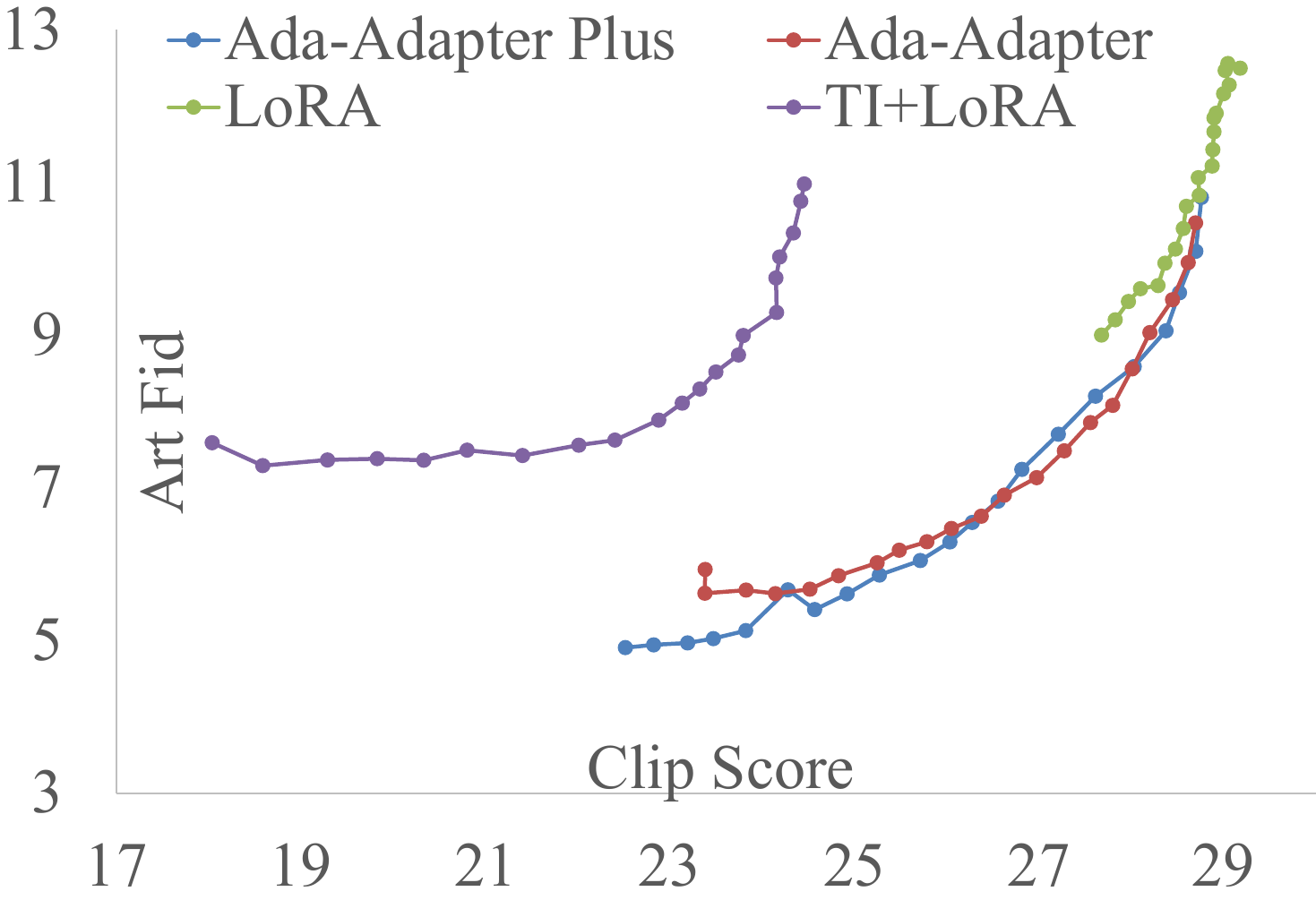}
\caption{Effect of various scale multipliers}
\label{fig:ablation_ip_scale}
\end{minipage}
\vspace{-0.3in}
\end{figure}

\section{Ablation study}
\subsection{The number of reference images}
To explore the potential of our method, we conduct experiments based on multiple datasets of various image counts and the evaluation results are shown in Fig.~\ref{fig:ablation_image_count}. In extreme cases, we only use 1 image to train Ada-Adapter, with the visualization results shown in Fig.~\ref{fig:ablation_one_shot}. We can conclude that our method is able to perform stable stylization even when only 1 image is provided for training.

\begin{table}[b]
\vspace{-0.1in}
\centering
\resizebox{0.6\linewidth}{!}{
\begin{tabular}{p{2.5cm}<{\centering}|p{2.5cm}<{\centering}|p{2.5cm}<{\centering}}
\toprule
            & W/o H-scales  & W/ H-scales \\
\midrule
Art Fid ($\downarrow$)        & \textbf{4.937}  & 5.187      \\                                        
Clip Score ($\uparrow$)        & 22.970  & \textbf{24.179}     \\                                                  
\bottomrule
\end{tabular}
}
\caption{The results of our pipeline with and without hierarchical scales. We use H-scales to represent our hierarchical strategy in short.}
\label{tab:ablation_study_hierarchical_weights}
\vspace{-0.3in}
\end{table}



\vspace{-0.1in}
\subsection{The necessity for hierarchical scales}
\label{subsec:need_for_hierarchical_weights}
\begin{figure}
\vspace{-0.2in}
      \centering
  \includegraphics[width=0.7\linewidth]{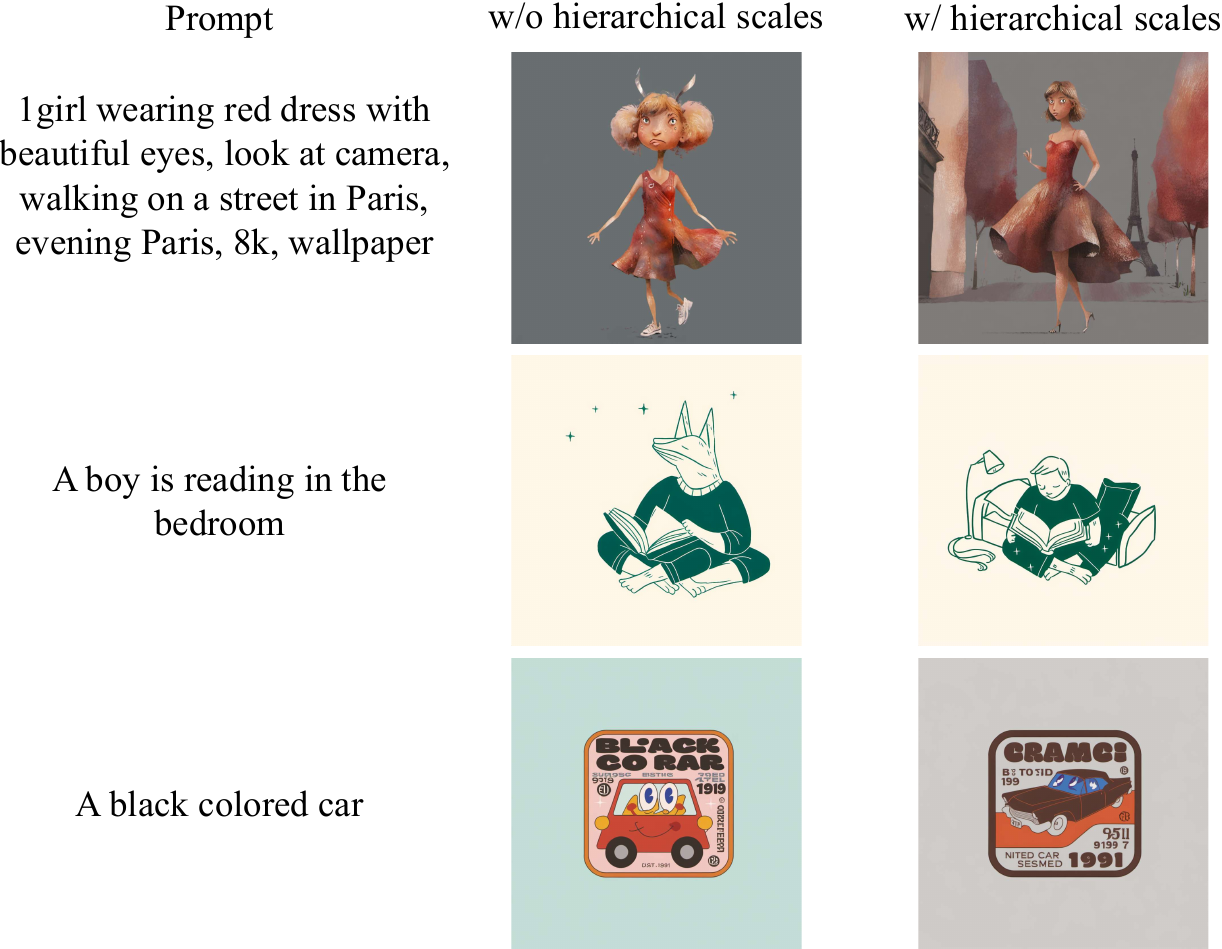}
  \caption{Result of Ada-Adapter pipeline with and without hierarchical scales. When applying hierarchical scales to our pipeline, the ability of text alignment notably grows, proving the importance of our hierarchical strategy.}
   \label{fig:ablation_with_or_without_hierarchical_weights}
\vspace{-0.4in}   
\end{figure}

We evaluate the impact of our hierarchical strategy by comparing our pipeline with and without the hierarchical scales for IP-Adapter. Fig.~\ref{fig:ablation_with_or_without_hierarchical_weights} shows the results. The stylized images with hierarchical scales have better alignment with the text prompts than those without hierarchical scales. Tab.~\ref{tab:ablation_study_hierarchical_weights} shows the evaluation metrics for this comparison. The Art Fid slightly drops, but the Clip Score significantly improves when we use hierarchical scales. This suggests that the hierarchical scales help to separate style and subject features and enhance text alignment, with a only small trade-off in style fidelity.

\subsection{Effect of scale multiplier at inference time}
\label{subsec:ab_scale_multiplier}

In our pipeline, the image encoders from IP-Adapter~\cite{ye2023ipadapter} provides abundant style prior knowledge, and the scale of image-conditioned features affects the generation of stylized images. Therefore, we bring in a hyper-parameter called adapter scale multiplier which directly multiplies with features of image encoders after we set scale for each layer at training time. For instance, if we set scale to 0.8 for a cross attention layer of IP-Adapter during training, and set adapter scale multiplier to 0.5 while performing inference, the final scaling factor for image features of IP-Adapter will be $0.8 \times 0.5 = 0.4$.
\begin{figure}
\vspace{-0.2in}
  \centering
  \includegraphics[width=1\linewidth]{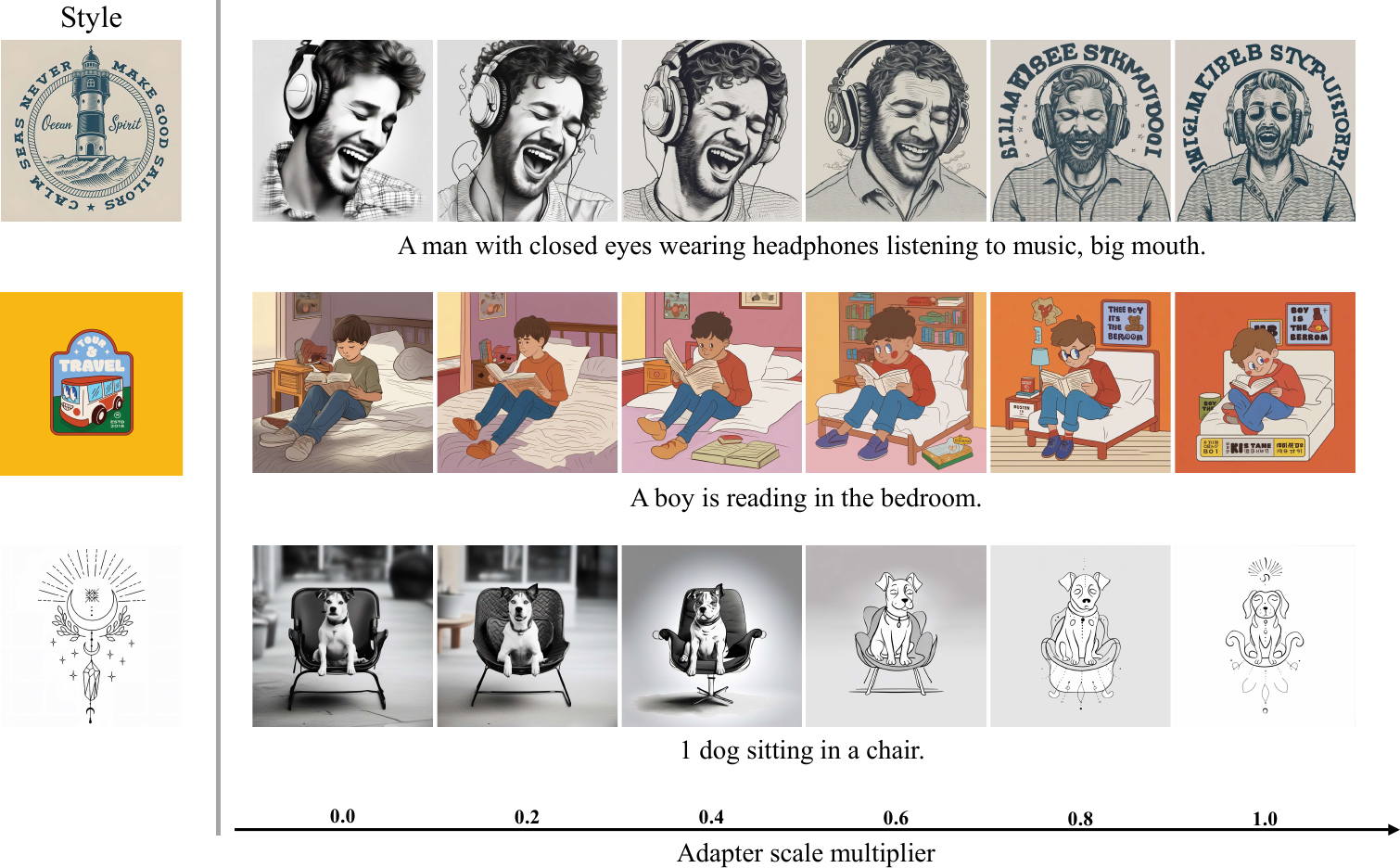}
  \caption{The effect of adapter scale multiplier on the stylization quality.}
   \label{fig:ablation_ip_scale_vis}
\vspace{-0.2in}   
\end{figure}
Similarly, for vanilla LoRA and TI + LoRA, we modify their function by adjusting LoRA scale multipliers.

We investigate how different scale multipliers affect the Art Fid and Clip Score of the stylized images in Fig.~\ref{fig:ablation_ip_scale}. We incrementally adjust the scale multiplier from 0.0 to 1.0 in steps of 0.05, plotting the corresponding trends of Art Fid and Clip Score. Each point in the figure represents the evaluation result of a specific scale multiplier. The lower-right corner of the figure indicates the best style and text alignment. 
Unlike LoRA and TI+LoRA, which struggle to simultaneously satisfy semantic and stylistic accuracy, our methods demonstrate superior performance.

We visualize some points in Fig.~\ref{fig:ablation_ip_scale} and show the results in Fig.~\ref{fig:ablation_ip_scale_vis}. The stylization is performed when a relatively large scale multiplier is applied to IP-Adapter, and the stylized images become closer to the datasets when the adapter scale multiplier increases.

\section{Conclusion}
Our method, Ada-Adapter, leverages a pre-trained image encoder to extract style features and significantly lowers the cost of style personalization using off-the-shelf diffusion models. 
Requiring merely a handful of source style images and a brief fine-tuning period, Ada-Adapter facilitates stable and consistent stylization while retaining the diffusion models’ text comprehension capabilities. Our method not only streamlines the artistic creation process, enabling artists to produce works with personalized styles swiftly, but also ensures ease of integration into existing workflows, offering a practical and efficient solution for both individual creators and the creative industry at large.

%
%
\bibliographystyle{splncs04}
\bibliography{egbib}
\end{document}